\documentclass[12pt]{article}

\usepackage[letterpaper, margin=1in]{geometry}
\usepackage{setspace}
\usepackage{authblk}
\usepackage{wrapfig}
\usepackage{graphicx}
\usepackage{comment}
\usepackage{enumerate}
\usepackage{mathtools}

\usepackage[pdftex,pagebackref=false]{hyperref} 
\hypersetup{
    plainpages=false,       
    unicode=false,          
    pdftoolbar=true,        
    pdfmenubar=true,        
    pdffitwindow=false,     
    pdfstartview={FitH},    
    pdfnewwindow=true,      
    colorlinks=true,        
    linkcolor=blue,         
    citecolor=green,        
    filecolor=magenta,      
    urlcolor=cyan           
}

\title{\vspace{-1.0cm}Investigating the Automatic Classification of Algae Using Fusion of Spectral and Morphological Characteristics of Algae via Deep Residual Learning}

\author[1,2]{\textbf{Jason L. Deglint}}
\author[1,3]{\textbf{Chao Jin}}
\author[1,2]{\textbf{Alexander Wong}}

\affil[1]{{\scriptsize Vision and Imaging Processing (VIP) Lab, Department of Systems Design Engineering, \newline University of Waterloo, Waterloo, Ontario, Canada}}
\affil[2]{{\scriptsize Waterloo Artificial Intelligence Institute, Waterloo, Ontario, Canada}}
\affil[3]{{\scriptsize School of Environmental Science and Engineering, Sun Yat-sen University, Guangzhou 510275, PR China}}

\date{October 3, 2018}

\begin{document}
\maketitle

\begin{abstract}

Under the impact of global climate changes and human activities, harmful algae blooms (HABs) in surface waters have become a growing concern due to negative impacts on water related industries, such as tourism, fishing and safe water supply. Many jurisdictions have introduced specific water quality regulations to protect public health and safety. Currently, drinking water quality guidelines related to cyanobacteria are based on maximum acceptable concentrations of toxins or elevated levels of cyanobacteria cells in water supplies. Therefore, reliable and cost effective methods of quantifying the type and concentration of threshold levels of algae cells has become critical for ensuring successful water management. In this work, we present SAMSON, an innovative system to automatically classify multiple types of algae from different phyla groups by combining standard morphological features with their multi-wavelength signals. Two phyla with focused investigation in this study are the Cyanophyta phylum (blue-green algae), and the Chlorophyta phylum (green algae). 
To accomplish this, we use a custom-designed microscopy imaging system which is configured to image water samples at two fluorescent wavelengths and seven absorption wavelengths using discrete-wavelength high-powered light emitting diodes (LEDs). 
Powered by computer vision and machine learning, we investigate the possibility and effectiveness of automatic classification using a deep residual convolutional neural network. 
More specifically, a classification accuracy of 96\% was achieved in an experiment conducted with six different algae types. 
This high level of accuracy was achieved using a deep residual convolutional neural network that learns the optimal combination of spectral and morphological features. 
These findings elude to the possibility of leveraging a unique fingerprint of algae cell (i.e. spectral wavelengths and morphological features) to automatically distinguish different algae types. 
Our work herein demonstrates that, when coupled with multi-band fluorescence microscopy, machine learning algorithms can potentially be used as a robust and cost-effective tool for identifying and enumerating algae cells.
\end{abstract}

\newpage
\tableofcontents
\newpage

\doublespacing
\section{Introduction}

\begin{figure}[b!]
	\centering
	\includegraphics[width=\columnwidth]{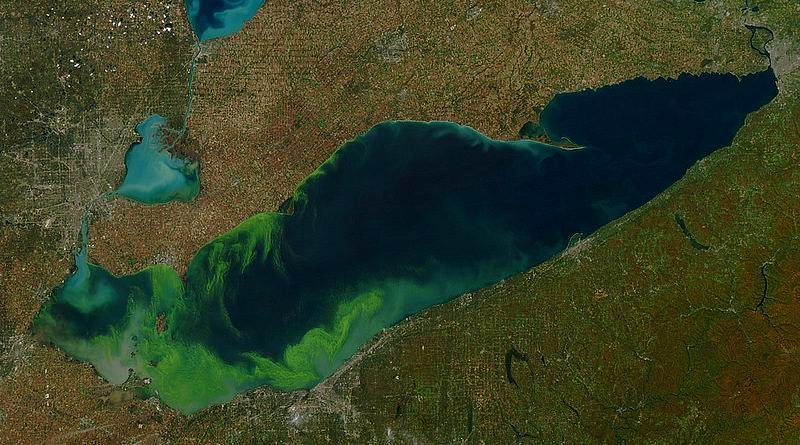}
	\caption{The Moderate Resolution Imaging Spectroradiometer (MODIS) on the Aqua satellite showing Lake Erie on October 9, 2011.  The bloom was primarily \textit{Microcystis Aeruginosa}, according to the Great Lakes Environmental Research Laboratory, which is a common type of cyanobacteria~\cite{nasa_2011}.}
	\label{fig:lakeErie}
\end{figure}

In the summer of 2011 Lake Erie experienced the largest harmful algae bloom (HAB) in recorded history~\cite{michalak2013record}.
As seen in Figure~\ref{fig:lakeErie}, this bloom was primarily \textit{Microcystis aeruginosa}, a type of algae which is one of the most lethal type of cyanobacteria, according to the Great Lakes Environmental Research Laboratory~\cite{nasa_2011}. 
Cyanobacteria can be extremely dangerous for humans and animals, as for example, swallowing \textit{Microcystis} can have serious side effects such as abdominal pain, diarrhea, vomiting, blistered mouths, dry coughs, and headaches.  
In addition, \textit{Anabaena}, another common cyanobacteria, can produce lethal neurotoxins called anatoxin-a which has shown to cause death by progressive respiratory paralysis~\cite{falconer1996potential}.

One toxin produced by \textit{Microcystis}, called microcystin-LR, is strictly regulated by the World Health Organization (WHO) as it is lethal for humans~\cite{world1998cyanobacterial}.
The maximum acceptable concentration (MAC) for the cyanobacteria toxin microcystin-LR in drinking water is 0.0015 mg/L (1.5 $\mu$g/L), according to the Government of Canada ~\cite{2002HealthCanada, guidelines_2012_healthCanada}.
In addition, in 2014 the U.S.A. released the Harmful Algal Bloom and Hypoxia Research and Control Amendments Act (HABHRCA) which requires the National Oceanic and Atmospheric Administration (NOAA) and United States Environmental Protection Agency (USEPA) to advance the scientific understanding and ability to detect, monitor, assess, and predict HABs and hypoxia events in marine and freshwater in the United States~\cite{congress2014harmful}. 
Therefore monitoring of water quality for different cyanobacteria and other micro-algae is essential for the proper management of any water body~\cite{coltelli2014water}.  The preservation and maintenance of our water directly affects marine wildlife, as well as the recreational, fishing and tourism industries, and most importantly water treatments plants that ensure clean drinking water is distributed to the population.

\section{Related Work}

On the on-set of bloom or during a bloom it can still be relatively challenging to determine which species of algae are present.
We will discuss three common methods to determine the abidance's of different algae types. 
In this paper we will discuss manual identification (Section~\ref{sec:manual}), using fluorescent probes (Section~\ref{sec:probes}), and imaging flow cytometry (Section~\ref{sec:IFC}).

\subsection{Manual Identification} \label{sec:manual}

The standard method of identifying and enumerating microalgae consists of three main steps which are: (1) sample preparation, (2) classification, and (3) enumerating.  
This current method of manual identification and enumeration by a taxonomist via a microscope is time consuming and very tedious.
Furthermore, each taxonomist needs years of specialised training and extensive experience to classify algae adequately~\cite{coltelli2014water}.  
Clerck \textit{et al.} presented their findings that the number of species of algae taxonomists are decreasing each year, resulting in more species needed to be identified by each taxonomist~\cite{clerck2013algal}.
They also show that presently we know of around 3000 algae species, while by the year 2200 an estimate of 6000 algae species will be known.
These trends indicate that it will become even more challenging for taxonomists to quickly and reliable classify various types of microalgae.
Finally, work presented by Culverhouse \textit{et al.} shows that human taxonomists have classification hit rate between 67\% and 83\%, depending on the taxonomist~\cite{culverhouse2003experts}.  This study shows that these experts are not unanimous in their classification, even when inspecting organisms with very similar distinct morphology~\cite{culverhouse2003experts, sieracki2010optical, colares2013microalgae}.
Unfortunately,  the current method of algae identification is quickly becoming unsustainable.

\subsection{Fluorescent Probes} \label{sec:probes}

A number of probes have been developed and are currently on the market that leverage the auto-fluorescence of algae.
For example, McQuaid \textit{et al.} used a YSI 660 V2-4 water quality multi-probe in their experiments, which was designed to measure the cyanobacteria’s phycocyanin pigment at 590 nm (with a passband of 565 nm - 605 nm) and measures the pigment’s emission at 660 nm $\pm$ 20 nm~\cite{mcquaid2011use}.  
They found that this probe is best used to monitor cyanobacterial biovolume in surface water when the cyanobacterial blooms were dominated by \textit{Microcystis sp.} and microcystin.
In 2012 Zamyadi \textit{et al.} took five different probes, some of which were YSI probes, and found that there was no correlation between a given probe's reading and the true cell count in a given sample~\cite{zamyadi2012cyanobacterial}.
However, the authors did find that the correlation between the probe's readings and the total biovolume in the sample could be trusted.
Zamyadi \textit{et al.} continued their work and in 2016 tested six different in-situ fluorometric probes from major brands such as bbe, TriOS, Tuner Designs, and YSI~\cite{zamyadi2016review}.
Finally, Bowling \textit{et al.} also used a YSI EXO2 fluorometric probe to measure the chlorophyll \textit{a} and phycocyanin and found that a good correlation between phycocyanin and total cyanobacterial biovolume in two of the three ponds they investigated~\cite{bowling2016assessment}.
They also found that phycocyanin did not correlate well with cell counts, and that Chl-\textit{a} was a poor measure of cyanobacterial presence.

In summary, probes that measure the Chl-a and phycocyanin can estimate the total biovolume in a cyanobacteria bloom, but are very poor at estimating actual cell counts of different species of cyanobacteria in the samples.
Another major disadvantage of all these probes is the fact that they cannot distinguish between species of cyanobacteria, as a  microscope would be required to accomplish this task.
Furthermore, a taxonomist analysis is shown to underestimate the risk of a microcystin contamination, due to less frequent sampling.
The major advantage of such probes is the ability of real-time data and automatic frequent sampling (at least every 60 minutes)~\cite{mcquaid2011use, zamyadi2012cyanobacterial, zamyadi2016review, bowling2016assessment, gregor2007detection}.

\subsection{Imaging Flow Cytometry} \label{sec:IFC}

The final method discussed in this paper to identify algae within water samples is to use an imaging flow cytometer, one common type being the FlowCAM.
In 2016, Corr\^{e}a \textit{et al.} used supervised learning on FlowCAM data which consisted on an imbalanced dataset of 24 types of microalgae divided in 19 classes, where the best performance classifier had the score 98.2\%~\cite{correa2016supervised}.  
Using the FlowCAM or other imaging flow cytometer is an effective method for algae classification, however, it still relatively expensive and constrained to a laboratory environment and requires proper training.

\section{Proposed Solution}

\begin{figure}[t!]
    \centering
    \includegraphics[width=\linewidth]{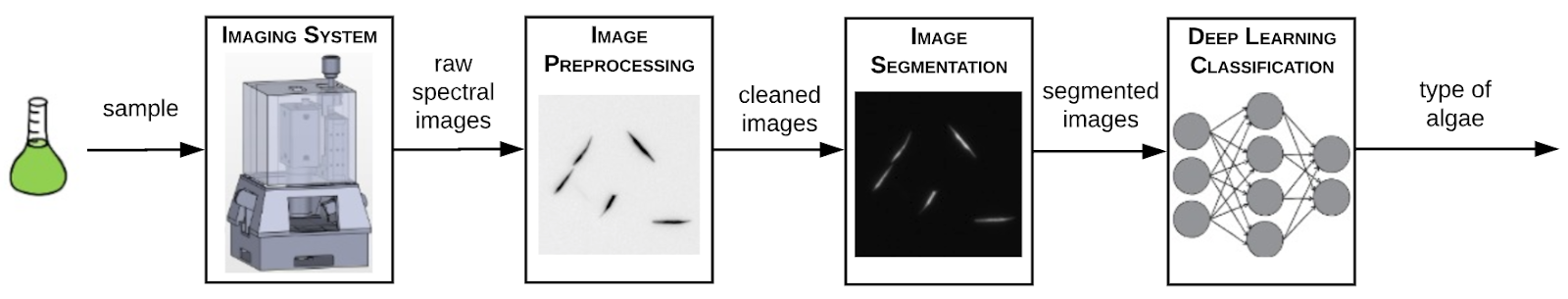}
    \caption{The SAMSON system is broken into four steps. First the water sample (Section~\ref{sec:algae}) are imaged using the imaging system (Section~\ref{sec:hardware}). This imaging data is then preprocessing (Section~\ref{sec:image_preprocessing}), and each organism is then segmented and cropped (Section~\ref{sec:image_segmentation}). Finally a deep residual learning-based image classification method is used to classify the algae type (Section~\ref{sec:dl}). }
    \label{fig:flow}
\end{figure}

The authors would like to propose an alternative method for the potential use of on-site water monitoring called \textbf{S}pectral \textbf{A}bsorption-fluorescence \textbf{M}icroscopy \textbf{S}ystem for \textbf{ON}-site-imaging (SAMSON) and is a continuation of SAMSON as initially presented by Deglint \textit{et al.}~\cite{CVIS2018}, where the authors presented their work of designing and building the imaging system.
In this work the authors would like to extend the capabilities of SAMSON and demonstrate it's capabilities by analysing the generated data using computer vision and machine learning techniques.
The extended SAMSON system, as shown in Figure~\ref{fig:flow}, is broken into four steps.
First the water sample (Section~\ref{sec:algae}) are imaged using the imaging system (Section~\ref{sec:hardware}). 
Together these two steps, or preparing the water sample and imaging it, make up the data collection component (Section~\ref{sec:datacolection}). 
Once having acquired the data, it can now be processed and analysed (Section~\ref{sec:data_analysis}). 
This can be broken into image preprocessing (Section~\ref{sec:image_preprocessing}), image segmentation (Section~\ref{sec:image_segmentation}), and finally deep residual learning-based image classification (Section~\ref{sec:dl}). 

\section{Data Collection} \label{sec:datacolection}

The data collection requires two steps.
First the algae sample  must be collected and prepared (Section~\ref{sec:algae}) and then a given water must be imaged using the SAMSON system (Section~\ref{sec:hardware}).

\subsection{Algae Samples} \label{sec:algae}

The two algae groups focused on in this research were the Chlorophyta phylum (green algae) and the Cyanophyta phylum (blue-green algae) since they are the most prevalent in harmful algae blooms.
As seen in Table~\ref{tab:algae}, certain pigments are contained in each phyla, such as chlorophyll-\textit{a} and chlorophyll-\textit{b} as well as $\beta$-carotene. 
However, blue-green algae are known to contain certain types of pigments that green algae do not contain, such as C-Phycoerythrin (CPE), C-Phycocyanin (CPC) and Allophycocyanin (APC).
This difference in pigmentation will later be leveraged for classification since these pigments occupy different parts of the electromagnetic spectrum and are known to absorb and fluoresce light differently~\cite{barsanti2014algae}.

\begin{table}[]
    \centering
    \caption{Different phyla of algae are known to have different pigments present in their cells~\cite{barsanti2014algae}. These difference or pigments are measured by SAMSON and used when building a machine learning algorithm to differentiate between types of algae.}
    \includegraphics[width=\linewidth]{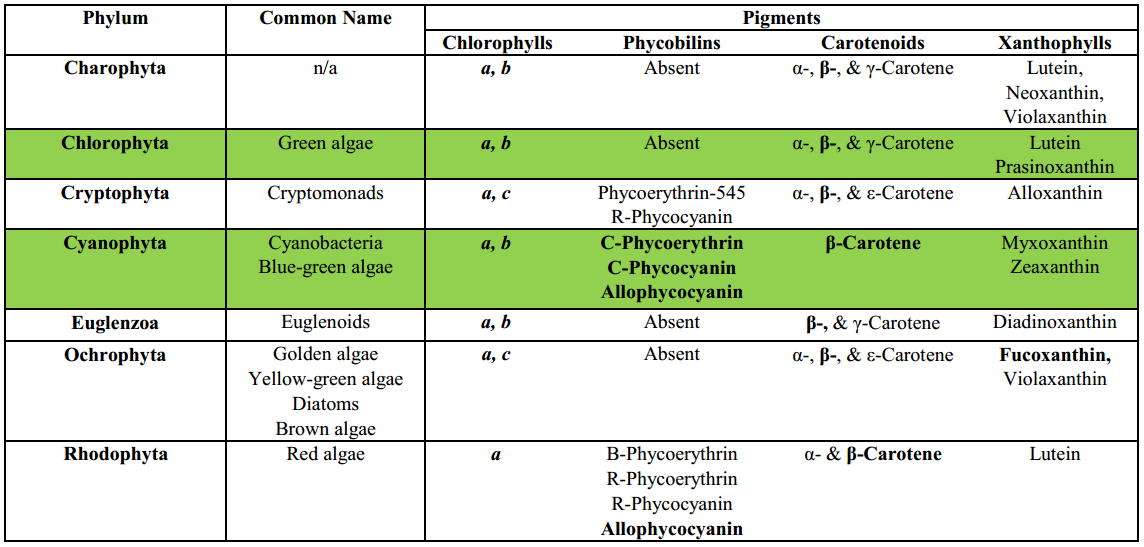}
    \label{tab:algae}
\end{table}

The six species of algae purchased from the Canadian Phycological Culture Centre (CPCC) were:
\begin{enumerate}[I.]
    \item Cyanophyta (blue-green algae or cyanobacteria)
    \begin{enumerate}[1.]
        \item \textit{Microcystis aeruginosa} (CPCC 300)
        \item \textit{Anabaena flos-aquae} (CPCC 067)
        \item \textit{Pseudanabaena tremula} (CPCC 471)
    \end{enumerate}
    \item Chlorophyta (green algae)
    \begin{enumerate}[1.]
    \setcounter{enumii}{3}
        \item \textit{Scenedesmus obliquus} (CPCC 005)
        \item \textit{Scenedesmus quadricauda} (CPCC 158)
        \item \textit{Ankistrodesmus falcatus} (CPCC 366)
    \end{enumerate}
\end{enumerate}

\textit{Microcystis aeruginosa} and \textit{Anabaena flos-aquae} where chosen as they are common culprits for producing toxins in a harmful algae bloom. 
\textit{Pseudanabaena tremula} was chosen since it is filamentous type of algae, just like \textit{Anabaena flos-aquae} and therefore may be difficult to distinguish the two types from each other.

\subsection{Imaging System} \label{sec:hardware}

\begin{figure}[]
    \centering
    \includegraphics[width=0.6\columnwidth]{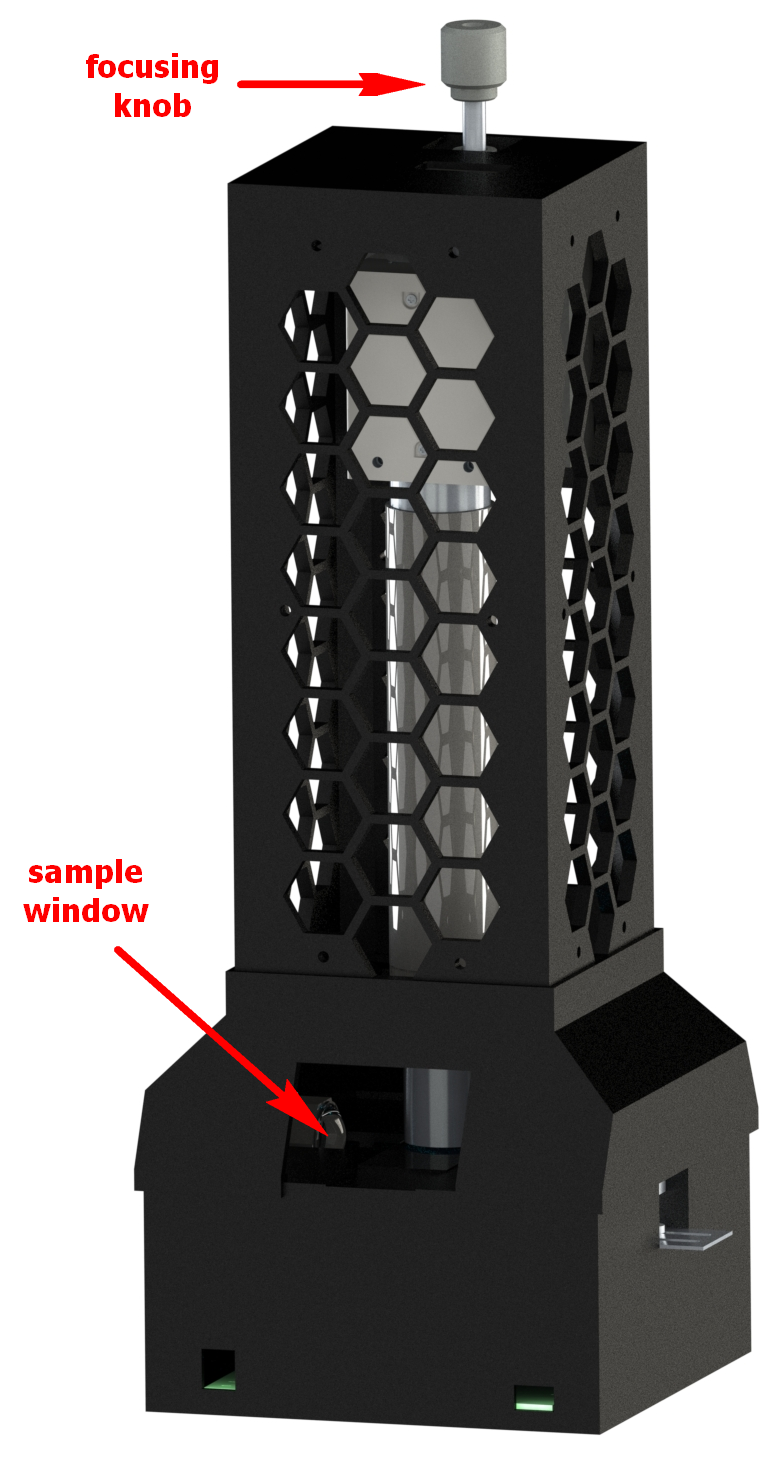}
    \caption{The SAMSON hardware system, initially presented by Deglint \textit{et al.}~\cite{CVIS2018} is used to collect multispectral images of water samples. In this work the authors collect two fluorescent images and seven absorption images, however SAMSON can be configured for a variety of different wavelength combinations. The user places the water sample slide in the slide window. Then using the graphical user interface (GUI) the user can view a live image of the sample and adjust the focus of the image using the focusing knob.}
    \label{fig:3d_print}
\end{figure}

A 3D render of SAMSON can be seen in Figure~\ref{fig:3d_print}, and has outer dimensions of 14 cm $\times$ 14 cm $\times$ by 40 cm.
This 3D model houses the scientific camera, the optics required to capture and focus the light, a slide holder for the water sample, as well as LEDs and a custom printed circuit board (PCB) to control the LEDs.
The nine LEDs chosen to image the six previously mentioned algae samples are:
\begin{enumerate}[I.]
\setlength{\itemsep}{0pt}
\setlength{\parskip}{0pt}
    \item Fluorescent LED wavelengths
    \begin{enumerate}[1.]
    \setlength{\itemsep}{0pt}
    \setlength{\parskip}{0pt}
        \item 385 nm (ultraviolet)
        \item 405 nm (ultraviolet)
    \end{enumerate}
    \item Absorption LED wavelengths:
    \begin{enumerate}[1.]
    \setlength{\itemsep}{0pt}
    \setlength{\parskip}{0pt}
        \item 465 nm (blue)
        \item 500 nm (cyan)
        \item 520 nm (green)
        \item 595 nm (amber)
        \item 620 nm (red-orange)
        \item 635 nm (red)
        \item 660 nm (deep-red)
    \end{enumerate}
\end{enumerate}

A custom user interface was also developed to control all the LEDs and and camera settings in order to capture the most relevant data. For full details on the hardware setup and graphical user interface please see work by Deglint \textit{et al.}~\cite{CVIS2018}.

\section{Data Analysis} \label{sec:data_analysis}

Having collected all the data it must now be preprocessed (Section~\ref{sec:image_preprocessing}), and then each organism in each image must be segmented and cropped (Section~\ref{sec:image_segmentation}).
Finally this new data can be used to construct a deep residual learning-based image classification system for classifying algae type (Section~\ref{sec:dl}). 

\subsection{Imaging Preprocessing} \label{sec:image_preprocessing}
\begin{figure}[]
    \centering
    \includegraphics[width=\linewidth]{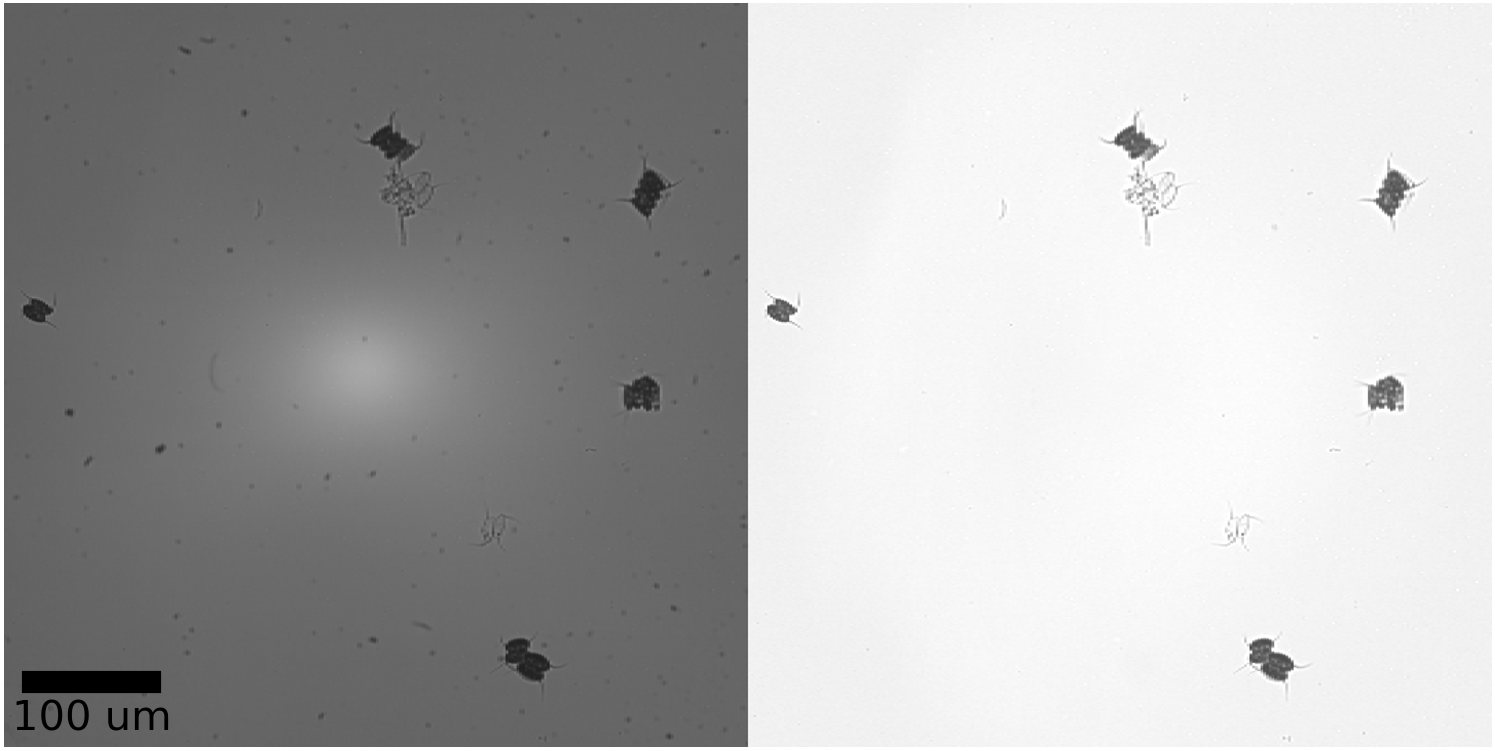}
    \caption{Using flat field correction (Section~\ref{sec:image_preprocessing}), the raw image (left) from the hardware system is then corrected (right). The result of flat field correction makes the task of image segmentation (Section~\ref{sec:image_segmentation}) much easier.}
    \label{fig:ffc}
\end{figure}

The first step in cleaning and preparing the data for a machine learning algorithm is to remove any background illumination inconsistencies, which can be accomplished by a method known as flat field correction~\cite{murphy2002fundamentals}.
Flat field correction can be mathematically described as
\begin{equation}
    I_C = \frac{I_R-I_D}{I_F-I_D}
\end{equation}
where $I_R$ is the raw image, $I_D$ is an image captured with no light source, that is a dark image, $I_F$ is a image with no sample and only the light source and $I_C$ is the corrected image.
In Figure~\ref{fig:ffc}, the raw image $I_R$ can be seen on the left and the corrected image $I_C$ can be seen on the right.
From Figure~\ref{fig:ffc} (left) one can observe the non-uniformity of the light  as there is a noticeable bright spot in the centre. 
After flat-field correction, as in Figure~\ref{fig:ffc} (right), the corrected image has a complete uniform background.
The other major benefit of flat-field correction is that is removes any other background artefacts, such as dust or impurities on the optical elements or camera sensor. 
This flat-field correction was applied to each absorption wavelength image for a given set of multi-band fluorescence absorption images.

\subsection{Imaging Segmentation} \label{sec:image_segmentation}
\begin{figure}[]
    \centering
    \includegraphics[width=\linewidth]{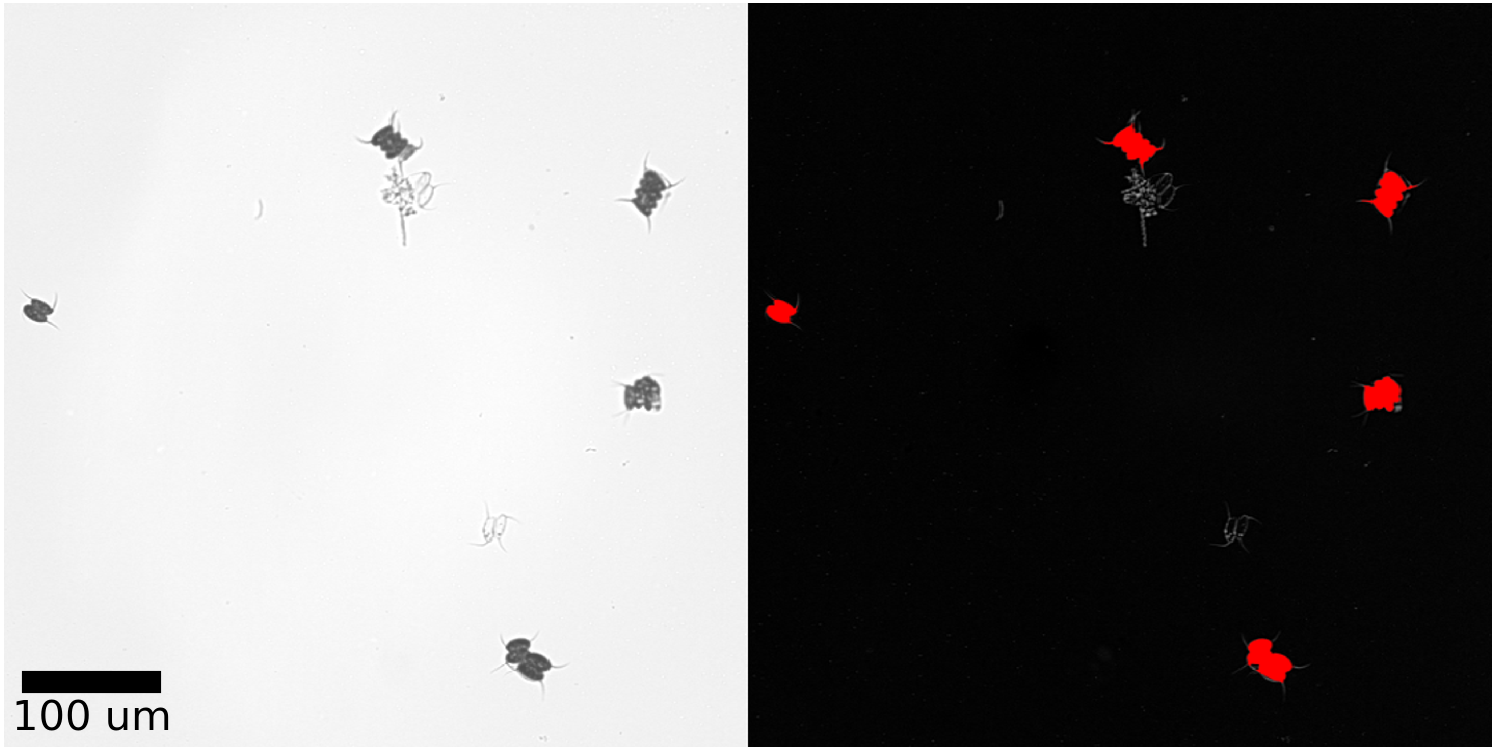}
    \caption{The preprocessed image (left) can be used to find an optimal threshold between the the foreground objects (algae) and the background. This threshold generates the segmented image (right) which can be used to locate and crop certain organisms. In this example the algae (red) have been segmented from the background.}
    \label{fig:segment}
\end{figure}

Given a corrected image the next challenge is to separate the background from the foreground as the algae samples are considered to be foreground objects.
Therefore a binary classifier was defined to classify each pixel into either the foreground class, $C_f$ or the background class, $C_b$.
The decision boundary of this classifier, $\theta$, was learned by implementing Otsu's method~\cite{otsu1975threshold}, where the inter-class variability of the image is maximised, which simultaneously minimises the intra-class variability. 
For any given pixel $\underline{x}$ the class, $C(\underline{x})$, was determined by:
\begin{equation}
C(\underline{x}) =
\begin{cases}
    C_f  & \text{if } f(\underline{x}) > \theta	\\
    C_b  & \text{otherwise}\\
  \end{cases}
\label{eq:binaryClassifier}
\end{equation}
\noindent where $f(\underline{x})$ is the pixel intensity at pixel $\underline{x}$.
As seen in Figure~\ref{fig:segment} (left) each pixel in $I_C$ is passed through the classifier, which results in the algae samples being segmented, as seen in Figure~\ref{fig:segment} (right).

\begin{figure}[t!]
    \centering
    \includegraphics[width=\linewidth]{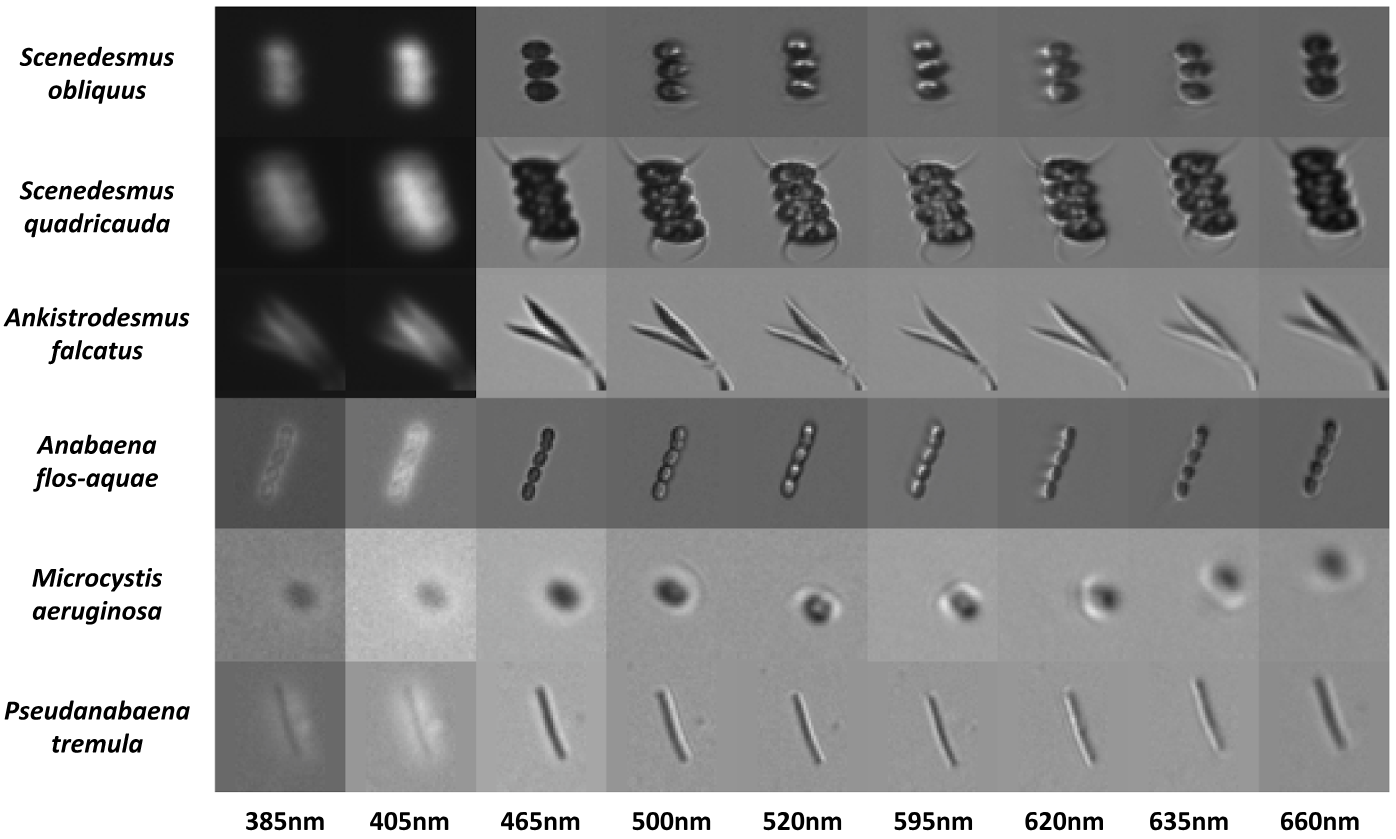}
    \caption{Six algae types were imaged at two fluorescent wavelengths (385 nm and 405 nm) as well as seven absorption wavelengths (465 nm, 500 nm, 520 nm, 595 nm, 620 nm, 635 nm, and 660 nm). Three of these algae are from the Cyanophyta phylum (blue-green algae) and the remaining three are from the Chlorophyta (green algae) phylum. These images are the result of segmenting and cropping the raw images from the hardware system.}
    \label{fig:cropped_images}
\end{figure}

Once all the organisms in a given multispectral image are segmented, each foreground group of pixels in the image were extracted and cropped to a fixed size. 
A sample cropped region of interest for each of the six species can be seen in Figure~\ref{fig:cropped_images}.
One initial observation is that all three of the green algae species have a much larger fluorescence signal at 385 nm and 405 nm compared to the blue-green algae, which matches results presented by Poryvkina \textit{et al.} findings~\cite{poryvkina2000analysis}. 
This difference in fluorescent intensity is due to the difference in pigmentation between each phylum, as previously discussed in Section~\ref{sec:algae}.

\begin{figure}[]
    \centering
    \includegraphics[width=\linewidth]{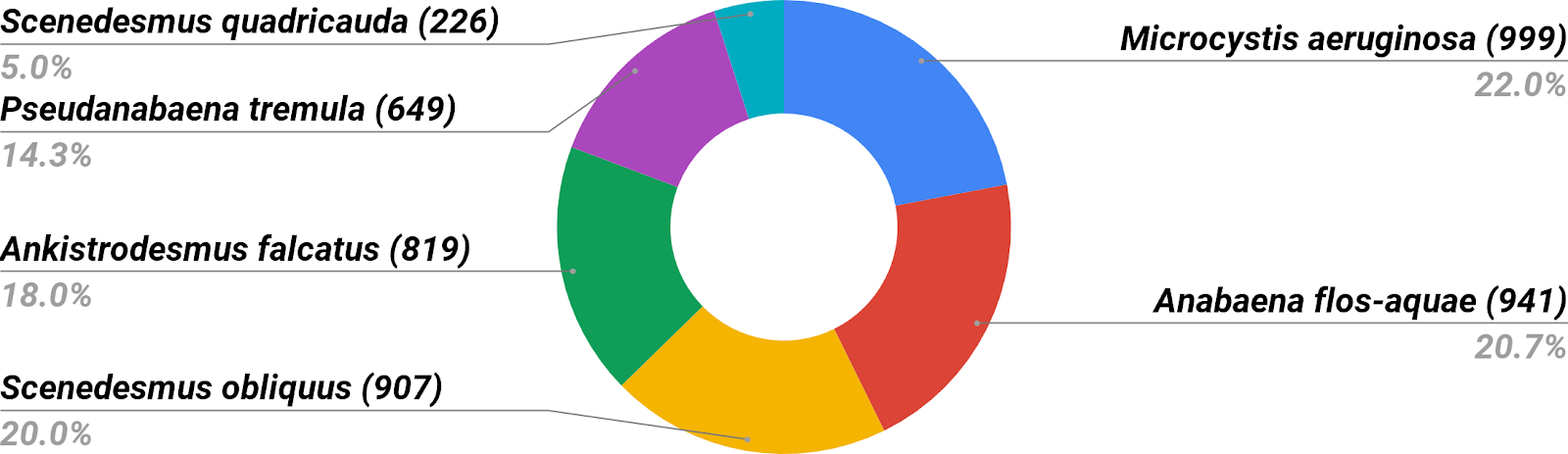}
    \caption{A total of 4541 segmented and cropped multispectral images were generated from the raw image collected from the imaging system. The class distribution of six types of algae can be seen above.}
    \label{fig:distribution}
\end{figure}

Each cropped image was then resized to a fixed dimension as requirement as input to the deep convolutional neural network, which results in the images losing their relative scale information.
For example, \textit{Microcystis aeruginosa} will appear larger than in the original image and the \textit{Anabaena flos-aquae} will appear smaller.
The distribution of how many cropped and resized images for each algae class can be seen in Figure~\ref{fig:distribution}.
The total number of multispectral images were 4541, that is, each of these 4541 images are composed of nine sub-images, two of which are fluorescence based, and seven which are absorption based.
This set of images makes up the available data to now train and test a deep neural network classifier.

\subsection{Deep Residual Learning-based Classification} \label{sec:dl}

The automatic classification of different types of algae was achieved via deep learning, which has been demonstrated in recent years to provide state-of-the-art performance across a wide variety of applications.  In particular, we leverage the concept of deep convolutional neural networks, a type of deep neural network in the realm of deep learning that has been demonstrated to be particularly effective for visual perception and understanding.  Here, we construct a custom 18-layer deep residual convolutional neural network that takes the captured multi-spectral image data as input, and outputs the predicted algae type.  A deep residual network architecture~\cite{he2016deep} was leveraged for its modeling capacity.  Due to the relatively small amount of data available, we leverage the concept of transfer learning when training this deep residual convolutional neural neural network, where the network is first trained on a larger dataset from a different domain prior to being finetuned for the task at hand.  This enables the network to build a strong mental model for characterizing image properties before being trained specifically to differentiate between different algae types.  More specifically, the deep residual convolutional neural network is first trained using the ImageNet dataset, a dataset of 1000 image classes containing over 14 million images.  After this training process, the network is then fine-tuned with 70\% of our available data.  Using the remaining 30\% of the available data to test the performance of the constructed network, it was found that the custom deep residual convolutional neural network was able to achieve a classification accuracy of 96\% (that is, 96\% of the data was classified correctly, while only 4\% was misidentified).

\begin{figure}[]
    \centering
    \includegraphics[width=0.55\linewidth]{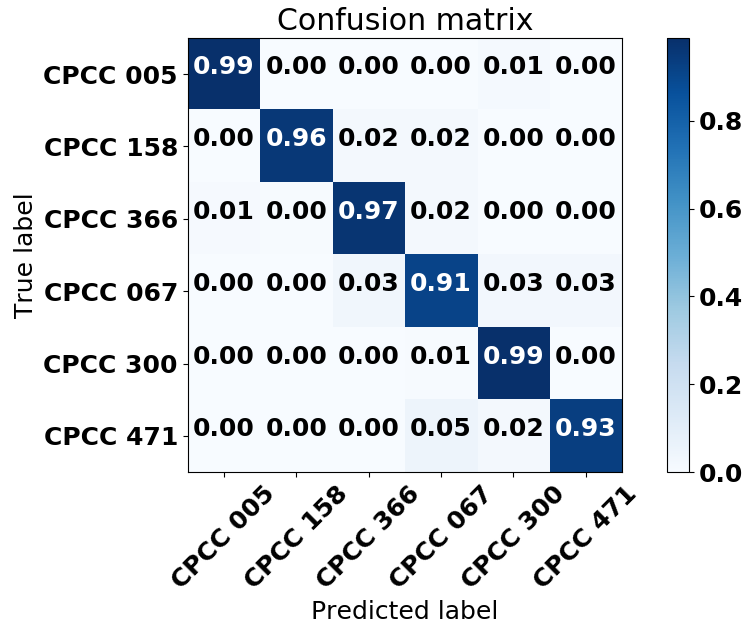}
    \caption{The confusion matrix is used to investigate the performance of the constructed deep residual convolutional neural network when classifying six types of algae. The overall classification accuracy of the constructed network is 96\%. The highest performing classes were were \textit{Scenedesmus obliquus} (CPCC 005) and \textit{Microcystis aeruginosa} (CPCC 300). }
    \label{fig:confusion}
\end{figure}

A confusion matrix, as seen in Figure~\ref{fig:confusion} was created to get a more nuanced understanding of the performance of the constructed deep residual convolutional neural network.
On the vertical axis of the confusion matrix we can see the true algae type for a given sample, while on the horizontal axis we see the predicted algae type.
For example, for CPCC 005 (\textit{Scenedesmus obliquus}), 99\% was classified correctly as CPCC 005, while 1\% was classified at CPCC 300 (\textit{Microcystis aeruginosa}).
Therefore the two highest performing classes were CPCC 005 (\textit{Scenedesmus obliquus}) and CPCC 300 (\textit{Microcystis aeruginosa}) each having a classification accuracy of 99\%.
The lowest classification accuracy was CPCC 067 (\textit{Anabaena flos-aquae}), as 3\% were miss-classified as CPCC 366 (\textit{Ankistrodesmus falcatus}), 3\% were miss-classified as CPCC 300 (\textit{Microcystis aeruginosa}), and 3\% were miss-classified as CPCC 471 (\textit{Pseudanabaena tremula}).
However, in each case the classification accuracy is higher than the reported 67\% - 83\% accuracy achieved by a human taxonomist~\cite{culverhouse2003experts}, while at the fraction of the time and by using a low-cost instrument.
This high performance demonstrates the potential use of such a system such as SAMSON for on-site use of algae identification.

\section{Conclusions} \label{sec:conclusions}

Current methods to determine which types of algae are present in an harmful algae bloom are time-consuming and relatively costly. 
For example, in a best case scenario manual identification by a highly trained professional can take a couple of days.
An imaging flow cytometer is an alternative solution, but require more training and are more costly.
Other alternatives of on-site monitoring, such as fluorescent probes have shown to be good at estimating bio-volume, but ineffective at identifying specific types of algae in a water sample.
Therefore a cost-effective on-site tool that can quickly and accurately identify different types of algae and bacteria in a water sample is highly desired.
By using the SAMSON system for data collection and the custom deep residual convolutiona neural network, we were able to achieve an accuracy of 96\% when classifying six different types of algae, either from the blue-green phylum or the green algae phylum.
This end-to-end approach allows a multispectral image to be input to the deep learning model and the corresponding type of algae is identified.
Furthermore, the main advantage of this method is that is learns the optimal combination of spectral and spatial features, due to the manner in which deep learning methods operate.
These initial results show that using a combination of fluorescence and absorption spectral data, along with the morphological data is a potentially effective method for on-site identification and monitoring of algae in a water body.

\addcontentsline{toc}{section}{Contributions}
\section*{Contributions}
JLD, CJ, and AW conceived and designed the SAMSON system.
JLD collected the data, wrote the code and ran the experiments.
JLD and AW designed the deep convolutional neural network for classification.
JLD, CJ, and AW conducted the analysis of the experiments.
JLD wrote the manuscript and JLD, CJ, and AW edited the manuscript.

\addcontentsline{toc}{section}{Acknowledgements}
\section*{Acknowledgements}
The authors would like to thank Heather Roshen at the Canadian Phycological Culture Centre (CPCC) for preparing the algae samples and Velocity Science for providing tools and resources for proper data collection.
This research was funded by the Natural Sciences and Engineering Research Council of Canada (NSERC) and Canada Research Chairs program.

\singlespacing
\addcontentsline{toc}{section}{References}
\bibliographystyle{unsrt}
\bibliography{uw-ethesis}

\end{document}